\documentclass[10pt,letterpaper]{article}

\usepackage{cogsci}

\usepackage{amsmath}
\cogscifinalcopy % Uncomment this line for the final submission 

\usepackage{graphicx}
\usepackage{pslatex}
\usepackage{apacite}
\usepackage{booktabs}

\usepackage{float} % Roger Levy added this and changed figure/table
\title{Can Large Language Models Predict Associations Among Human Attitudes?}

\author{{\large \bf Ana Ma (yutongm1@asu.edu)} \\
  School of Social and Behavioral Sciences \\%, 1202 W. Johnson Street \\
  Arizona State University
  \AND {\large \bf Derek Powell (dmpowell@asu.edu)} \\
  School of Social and Behavioral Sciences \\%, 1202 W. Johnson Street \\
  Arizona State University
  }

\begin{document}

\maketitle

\begin{abstract}
% Human beliefs are often interrelated both within and across social domains. While prior work has shown that large language models (LLMs) can predict human attitudes based on other similar attitudes, no study has directly examined whether models can represent human belief dynamics across distinct domains. In Study 1, we collected 376 participants’ attitudes toward 64 statements spanning 20 themes and found an overall correlational pattern among most attitudes. In Study 2, we prompted GPT-4 on all possible attitude pairs and observed strong and positive associations between the model’s predictions and human attitude correlations, and that the alignment was influenced by the semantic similarity between statements. To further investigate this influence, in Study 3, we collected GPT-4's predictions on human responses using some most semantically similar attitudes and those semantically dissimilar but statistically predictive (derived from human response patterns). Results revealed that while surface similarity improves prediction accuracy, the model still generated meaningful social inferences between dissimilar attitudes. These results suggest LLMs may capture some latent structure of human belief systems, but the mechanisms underlying this ability remain an open question for future research.
Prior work has shown that large language models (LLMs) can predict human attitudes based on other attitudes, but this work has largely focused on predictions from highly similar and interrelated attitudes. In contrast, human attitudes are often strongly associated even across disparate and dissimilar topics. Using a novel dataset of human responses toward diverse attitude statements, we found that a frontier language model (GPT-4o) was able to recreate the pairwise correlations among individual attitudes and to predict individuals' attitudes from one another. Crucially, in an advance over prior work, we tested GPT-4o's ability to predict in the absence of surface-similarity between attitudes, finding that while surface similarity improves prediction accuracy, the model was still highly-capable of generating meaningful social inferences between dissimilar attitudes. Altogether, our findings indicate that LLMs capture crucial aspects of the deeper, latent structure of human belief systems.

\end{abstract}

% \section{General Formatting Instructions}

Whether about politics, ethics, science, sports, or the weather, people's beliefs and attitudes rarely stand alone. Instead, they are interrelated with one another, interwoven through cultural and social influences \cite{Converse1964, Hofstede2001}, and fit together as part of their intuitive theories for how the world works \cite{gerstenbergIntuitiveTheories2017, weismanTheorybasedExplanationIntervention2017, PowellEtAl2023Modeling}. The ability to appreciate and anticipate these interrelations is a crucial target of human social reasoning and inference, as one generally only need to learn about a subset of a person's views to infer a great deal about the rest. 
% And understanding others' perspectives can be key for navigating relationships. Drawing such inferences is key to knowing what is or is not appropriate or not to say to grandma, what argument might persuade a colleague, or what explanation might help a confused student overcome a misconception or learn a new concept. [if we can, citations for those points. And something else about the ability to do this or not do this]

% To be sure, these kinds of social inferences are not necessarily straightforward and, until recently, would have been the exclusive domain of human reasoning. Now however, large language models (LLMs) pretrained on massive amounts of text data and further finetuned for instruction-following have shown remarkable facilities in a wide variety of tasks \cite{}. 
Researchers have now begun examining whether large language models (LLMs) might be able to predict or mimic the attitudes of different groups or individuals \cite{hwangAligningLanguageModels2023, santurkarWhoseOpinionsLanguage2023}. Within the Artificial Intelligence and Natural Language Processing communities, this is sometimes likened to adopting the "persona" of group or individual users. This could be useful for aligning chat assistants to the cultural perspectives of their users \cite{ChoenniShutova2024SelfAlignment, PawarEtAl2024Survey}, for developing automated educational tools \cite{RazafinirinaEtAl2024Pedagogical, SonkarEtAl2024Pedagogical, TesslerEtAl2024AI}, for targeting recommendations or advertising \cite{El-SayedEtAl2024MechanismBased, TangEtAl2024GenAI} \footnote{These uses present substantial risks and potential ethical concerns, as discussed in the works we cite here.}, or even for generating simulated human responses for social science research \cite{argyleOutOneMany2023}.

A number of recent works have found that LLMs were able to predict human attitudes from demographic, ideological, and attitudinal measures \cite<e.g.>{argyleOutOneMany2023, santurkarWhoseOpinionsLanguage2023, hwangAligningLanguageModels2023, LongEtAl2024Aligning}. For instance, \citeA{hwangAligningLanguageModels2023} found that prompting LLMs with a person's responses for a subset of related survey questions improved its ability to predict their other survey responses as compared to prompting with the respondents' demographics or ideology alone.

However, it remains unclear the extent to which these predictive capacities demonstrate LLM's ability to engage in social reasoning about relationships between human attitudes and opinions. Thus far, these models have only been tested in contexts where they have access to information about people's opinions for matters highly similar to the opinions-to-be-predicted. Therefore, it is possible that prior successes are driven largely by relatively surface-level semantic similarities between the prompted and predicted statements.

Conversely, a striking feature of human attitudes are the often strong interrelations among ostensibly distant attitudes or beliefs. Perhaps most obviously, people's views about political, religious, or moral concerns tend to be organized by broader ideologies. For instance, in the U.S., conservatives tend to favor strong restrictions on abortion while favoring little to no restriction on gun sales. Meanwhile, many other beliefs are related due to their intrinsic interconnections. For instance, \citeA{PowellEtAl2023Modeling} examined intuitive theories surrounding vaccination decisions. They found that people's ``vaccination intentions'' were most strongly related to highly proximal beliefs, such as about ``vaccine danger'' and ``vaccine effectiveness''. However, they also found quite strong relationships between vaccination intentions and more distant beliefs, such as about the merits of natural versus artificial things, and about the role of ``balance'' in determining health.

As these examples illustrate, relations among people's beliefs and attitudes can be non-obvious---e.g., gun control and reproductive rights have nothing in common on their face---and therefore unlikely to be predictable from the semantic similarity of statements of the beliefs themselves. Rather, appreciating these interrelations requires a deeper understanding of both social and epistemic concerns. 

\subsection{Interrelations among human attitudes and beliefs}

% A person’s attitudes and opinions are rarely isolated and are often coherent amongst one another following certain patterns. 
A \textit{belief system} consists of a grouping of human attitudes that can form via a range of constraints \cite{Converse1964}. One constraint is ideology: for example, conservatism is characterized by resistance to change and a preference for tradition \cite{Adorno1950}, and is associated with both political attitudes (e.g., against legalized abortion) as well as non-political views \cite<e.g., skepticism of modern art;>{Wilson1968}. Moreover, ideologies themselves are frequently correlated with one another \cite{Jost2003Conservatism}. For example, correlations have been found among political conservatism, Social Dominance Orientation \cite<i.e., support for hierarchical social structures;>{Pratto1999, Altemeyer1998}, and System Justification Theory \cite<i.e., legitimizing existing systems;>{Jost2003}. Through the linkages among ideologies, an even broader web of attitudes can be drawn together into a system of beliefs. 

Social influences are another source of constraint on cohesive belief patterns. Social ties promote interactions that reward similarity and alignment in behaviors \cite{Schachter1959} and attitudes \cite{Byrne1961, Moussaid2013Social}. Individuals' come to internalize the social norms they experience across familial, educational, religious, and cultural environments, producing bundles of related beliefs \cite{Converse1964}. For example, a community with Individualistic norms can establish attitudes about work (e.g., valuing personal achievement over collective gain), romantic relationships (e.g., emphasizing personal choice over arranged relationships), philosophy of life (e.g., focusing on finding one’s own path over fulfilling a societal role), and other concerns \cite{Hofstede2001}. Not only are individuals passively influenced by these social norms, but they also further actively self-select into groups of shared views. This process, termed Homophily \cite{McPherson2001},  creates loops that further reinforce the groupings of attitudes \cite{axelrod2021preventing}.

Of course, perhaps the most fundamental constraint on beliefs and attitudes is \textit{coherence} \cite{thagardExplanatoryCoherence1989}: attitudes and beliefs are connected with one another through their meanings, through logical implication, and through people's understanding of causal relationships in the world \cite{gerstenbergIntuitiveTheories2017, PowellEtAl2023Modeling}. In the face of new evidence, people attempt to update their beliefs to maintain coherence \cite{holyoakBidirectionalReasoningDecision1999, spellmanCoherenceModelCognitive1993}, and a clash between beliefs can lead to the discomfort of ``cognitive dissonance'' \cite{festinger1957cognitive}. 

\subsection{Social inference and LLM personas}

% \citeA{santurkarWhoseOpinionsLanguage2023} leveraged a set of questions and survey responses collected by Pew to create \textsc{OpinionQA}, a benchmark dataset that they use to examine the alignment of several LLMs with human opinions across a wide range of demographic groups. Prompting LLMs to adopt the personas of these demographic subgroups, they found that while the models' predictions did better at capturing the group's attitudes than a model prompted without a persona, LLMs' predictions and actual human opinions remained largely misaligned. \citeA{hwangAligningLanguageModels2023} found greater success by prompting an LLM with individual respondent's responses for a subset of related opinion questions. A model prompted with the responses for the most-similar opinions provided by that respondent was substantially more accurate at predicting new responses than a model without this information.

\textsc{OpinionQA} \cite{santurkarWhoseOpinionsLanguage2023} has emerged as a leading benchmark for researchers and developers aiming to align LLMs to users along demographic, ideological, and attitudinal dimensions. This benchmark leverages high-quality survey data collected by Pew from several representative samples of U.S. respondents to provide both categorical and individual-level persona information. The benchmark is composed of 1506 survey questions and answers from 80,098 respondents measured across 15 American Trends panel surveys conducted by Pew. Each American Trends panel survey focuses on a set of topics of concern for American civil and political life.
% For instance, one panel focused on "Guns", and another focused on "Automation and driverless vehicles".

A number of prior works have focused on aligning LLM responses with specific groups of users \cite{santurkarWhoseOpinionsLanguage2023}. 
% However, this neglects the substantial variability of attitudes within groups, and the general nuance of human opinion and attitudes. 
Going further, \citeA{hwangAligningLanguageModels2023} examined more fine-grained relationships between specific attitudes. They found that prompting an LLM with a person's responses to other specific attitude questions substantially improved the model's prediction of their predictionr of a target attitude. Moreover, they found that selecting a subset of attitudes for prompting improved predictions: filtering to choose the top-k attitude questions most similar to the target statement (in terms of cosine similarity) improved predictions as compared with prompting with all responses.

Due to the structure of the Pew panel data on which the \textsc{OpinionQA} benchmark is based, survey respondents tended to be asked about a number of similar questions. As such, \citeA{hwangAligningLanguageModels2023}'s methodology for selecting the top-k most similar items and responses as the language model prompt will tend to produce highly-similar prompting sets, which might allow for successful prediction from a relatively simple inference processes. Roughly, good predictions might be achieved through a heuristic process of similarity-matching, something like: predict the response given to the most similar item in the prompt. Subsequent works have found that using a more sophisticated process to select prompt information can produce more accurate predictions \cite{LongEtAl2024Aligning}, suggesting that modern LLMs engage in inferential processing beyond simple similarity-matching heuristics. However, this work did not examined this question directly.

Another line of work has examined the potential application of LLMs in simulating human responses for social science research \cite{argyleOutOneMany2023, SunEtAl2024Random}. To validate the viability of this application, \citeA{argyleOutOneMany2023} define and evaluate several criteria for establishing algorithmic fidelity to human responses. Most important for our question is what they call ``Pattern Correspondence'': whether correlations among attitudes from silica samples reflect those in human samples.
% This reflects our question regarding social inference: can an LLM predict how attitudes relate? \citeA{argyleOutOneMany2023} examined this for 12 survey questions (6 demographic, 6 attitudinal) from the ANES 2016 survey. 
% Use GPT-3 \cite{} to predict each survey responses from a human respondents' answers for the other 11. Then, from this simulated data, they calculated a measure of association between each item. 
% They report that the simulated data successfully recapitulate the associations among variables observed among human participants. However, these 12 survey questions provide only a limited view into the attitudes represented by the model, and 
As in other work using \textsc{OpinionQA} \cite{hwangAligningLanguageModels2023, LongEtAl2024Aligning}, their findings may owe to semantic similarity of the items measured. 
% Of the 6 attitudes measured, 5 concern politics or political ideology, and, perhaps notably, the only non-political item (``attend church'') saw the largest discrepancies between human and model-generated associations. 

\subsection{The present studies}

We sought to examine the the extent to which frontier language models can predict peoples' attitudes from one another in the absence of direct semantic similarity between the prompted and target attitudes. To support this novel examination across more distant connections, we surveyed a sample of U.S. respondents on a wide-ranging subset opinion questions from the Pew surveys underlying \textsc{OpinionQA} \cite{santurkarWhoseOpinionsLanguage2023}. Then, we tested the ability of OpenAI's GPT-4o \cite{OpenAIEtAl2024GPT4o} to predict individual attitudes on the basis of their other survey responses, for both semantically-similar and semantically-dissimilar items.
% Prior work has examined LLMs ability to predict original survey responses from Pew. However, by construction, Pew surveys tend to target a single topic area (e.g. gun control), resulting in greater average similarity among attitude statements and preventing analysis of correlations among attitudes across topic areas or domains. 
% Asking a single set of respondents to answer a wide-ranging set of opinion questions supports novel examinations across more distant connections.
% Using this data, we examined the overall matrix of correlations among attitudes and the predictability of specific attitudes from others using a suite of models. We then tested the ability of OpenAI's GPT-4o \cite{OpenAIEtAl2024GPT4o} to predict individual attitudes on the basis of prior survey responses.
% , using several prompting strategies including a chain-of-thought prompting technique. 
% First, we prompted GPT-4o with individual example attitudes and examined its ability to recreate the pairwise correlations among individual attitudes. Then, we examined its ability to predict a target attitude from a subset of related attitudes, chosen either based on their semantic similarity, or on their observed correlations with the target attitude. 

Across our analyses, we find that GPT-4o imperfectly but meaningfully recreates the observed pairwise correlations among attitudes and predicted individual attitudes. Semantic similarity between attitude statements played a limited role in this capability: GPT-4o's predictions were biased by similarity and were more accurate when made based on semantically-similar items, but this influence was relatively modest. Our results clearly demonstrate that GPT-4o is able to make meaningful social inferences in the absence of surface-level semantic similarity between prompted and targeted attitude statements.
% Moreover, GPT-4o was able to predict people's attitudes just as accurately from dissimilar-but-relevant attitudes as it was from similar attitudes.
%Encouraging the model to engage in chain-of-thought reasoning improved performance [modestly/substantially]. 
Altogether, our findings indicate that frontier language models like GPT-4o can engage in social reasoning to predict individuals' attitudes.

\section{Study 1: Human Data Collection}

\begin{figure*}[!h]
    \begin{center}
    \includegraphics[width=.8\textwidth]{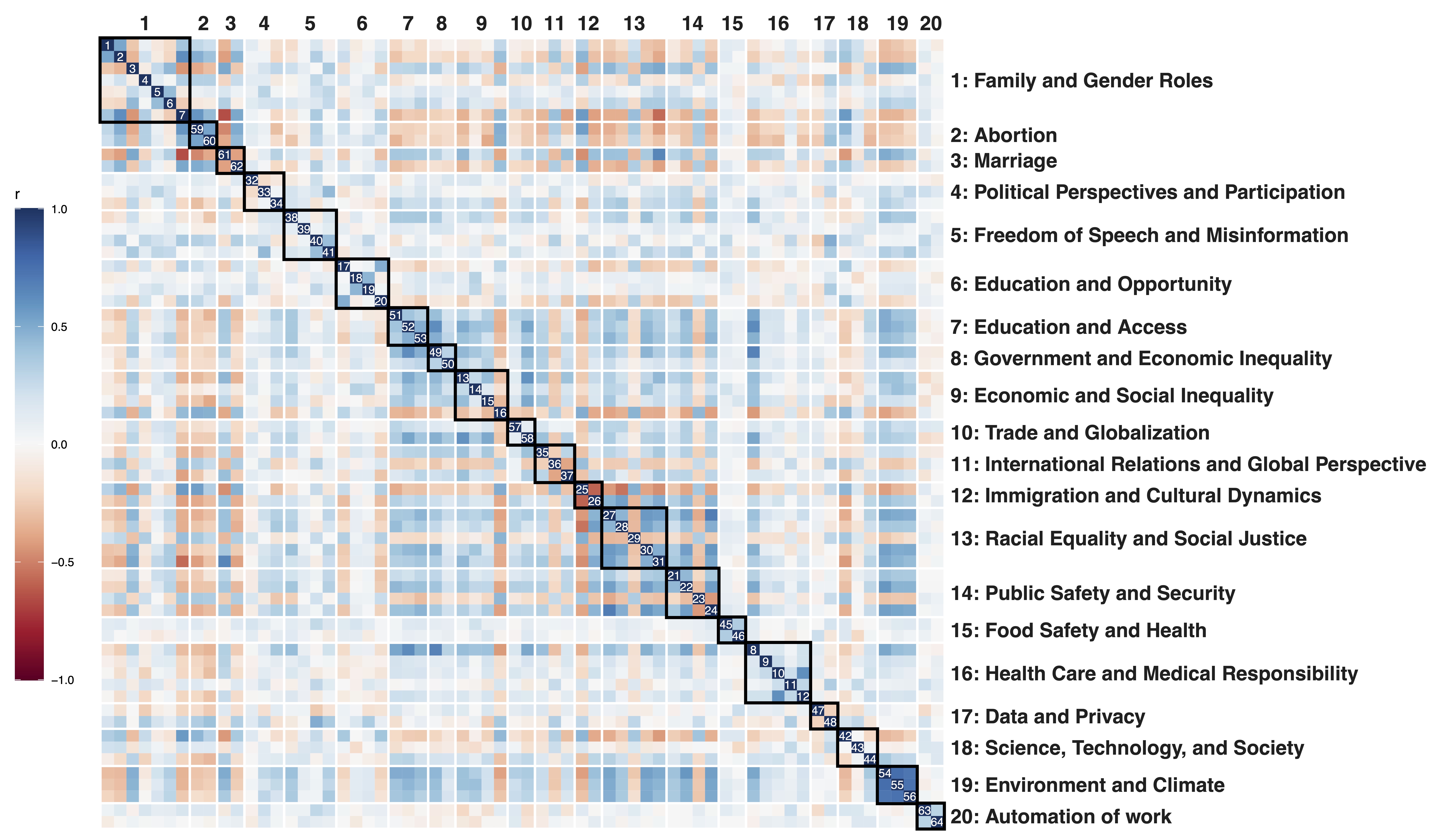}
    \end{center}
    \caption{Correlations among self-reported human attitudes. Attitudes (labeled along the diagonal) are ordered by topic area (labeled on axes), and topic areas are separated by grid lines. Black bounding boxes highlight topic groupings along the diagonal indicating within-topic associations.}
    \label{fig:human attitude heatmap}
   % \textit{Note:} The full list of themes are: 1 - \textit{Family and Gender Roles}, 2 - \textit{Abortion}, 3 - \textit{Marriage}, 4 - \textit{Political Perspectives and Participation}, 5 - \textit{Freedom of Speech and Misinformation}, 6 - \textit{Education and Opportunity}, 7 - \textit{Education and Access}, 8 - \textit{Government and Economic Inequality}, 9 - \textit{Economic and Social Inequality}, 10 - \textit{Trade and Globalization}, 11 - \textit{International Relations and Global Perspective}, 12 - \textit{Immigration and Cultural Dynamics}, 13 - \textit{Racial Equality and Social Justice}, 14 - \textit{Public Safety and Security}, 15 - \textit{Food Safety and Health}, 16 - \textit{Health Care and Medical Responsibility}, 17 - \textit{Data and Privacy}, 18 - \textit{Science, Technology, and Society}, 19 - \textit{Environment and Climate}, and 20 - \textit{Automation of Work}.
\end{figure*}

% \subsection{Method}

\subsection{Participants}

A sample of U.S. adults were invited to participate in the survey via Connect with a compensation of \$2.25 for their time. After excluding 10 participants who failed a simple attention-check, 376 participants (223 male, 147 female, 4 non-binary; 18 to 73 years old: Avg. age = 37.41, SD = 11.47) responses were included in our study. Participants reported a variety of races (237 White or Caucasian, 72 Black or African American, 29 Asian, 17 Hispanic or Latinx, 14 Multiracial/Biracial, 1 Native American or Alaskan Native, 1 none of the listed above, and 5 preferred not to say) and political backgrounds (198 in the Democratic Party, 85 Republican, 80 Independent, 6 Other, and 7 preferred not to say). 

\subsection{Materials and Procedure}

Drawing on the opinion questions in \textsc{OpinionQA}, we selected a subset of 64 diverse items assessing views on a wide range of topics relevant to the U.S. society. We transformed these questions from Pew into declarative statements to allow responses on a common agree-disagree scale for all items.  

Participants were surveyed in an online Qualtrics survey. After consenting to participate, participants rated their agreement with each statement on a five-point scale, ranging from \textit{Strongly agree} to \textit{Strongly disagree}. All participants were asked to respond to all 64 statements in a random order. Two attention checks were evenly-spaced in the study.

\subsection{Results}

Pearson correlation coefficients were calculated between all statement pairs (n = 4,032). As shown in Figure \ref{fig:human attitude heatmap}, substantial associations were found among attitudes both within and across topic areas.
As an illustrative example, participants' agreement level between statements ``The government should prioritize addressing climate change.'' (\textit{Environment and Climate}) and ``Increasing the number of guns is bad for society.'' (\textit{Public Safety and Security}) were positively correlated $r = 0.58$. Many other correlated attitudes were found between distinct social topics, as represented by non-diagonal colored tiles in Figure \ref{fig:human attitude heatmap}.

\section{Study 2: Estimating correlations with GPT-4o}

% In Study 1, we found participants' social attitudes to be correlated not only within but also among social topics. In Study 2, we continued to investigate whether language models can also predict these correlation patterns across domains, as well as whether the correlations were associated with the statements' semantic similarity. 

Next, we tested whether predictions from GPT-4o reflect the correlational patterns observed between human responses to these 64 atttiude questions.

% \subsection{Method}

\subsection{Model prompting}

% In order to investigate whether LLM's response to the Social Attitude Survey show similar item inter-correlations as humans', we prompted a model to generate an agreement level to each question statement by referencing to a hypothetical person's response to another statement in the question pool. 

To prompt GPT-4o to make its predictions, we followed the general structure used by \citeA{hwangAligningLanguageModels2023} to create user prompts containing 1) instructions for the model to "Help predict a person's answers on a social attitudes survey"; 2) an example statement and answer; 3)  instructions to generate a prediction based on the example statement-answer pair; and 4) a target statement and answer choices for the model to predict (i.e., the five agreement levels). For every possible unique pairing of the 64 attitude statements we created five prompts, with one statement as the target and the other serving as the "example" with each of the 5 possible answer choices. This produced $64 \ \text{target statements} \times 63 \ \text{example statements} \times 5 \ \text{example answers}=20,160$ total prompts. We then passed these prompts to GPT-4o through the {OpenAI} API and recorded its responses.\footnote{In all cases we use version \texttt{gpt-4o-2024-08-06}, sampling up to 10 tokens with temperature 0.01 for consistency.} % All model responses followed the specified answer-only format. 
% Moreover, we calculated the weighted probability for each agreement level to ensure the probabilities of all levels for a statement sum to 1.

% \begin{figure}
%     \centering
%     \includegraphics[width=1\linewidth]{figures/example-prompt.png}
%     \caption{Example prompt to predict human response based on a statement-answer pair}
%     \label{fig:example prompt}
% \end{figure}

\subsection{Metrics}

%\subsubsection{GPT-4o survey responses}

%The final LLM response to each question statement was calculated by:

%\[
%\text{LLM response} = \sum_{i=1}^5 (v_i \cdot p_i)
%\]

%\noindent where $v_i$ is the rescaled agreement level as a numeric value from 0 (\textit{Strongly disagree}) to 1 (\textit{Strongly agree}), and $p_i$ is the probability of the agreement level.

\subsubsection{GPT-4o-estimated correlations}

The prompted agreement level and the corresponding GPT-4o-predict agreement levels were converted to a numeric score from 1 (\textit{Strongly disagree}) to 5 (\textit{Strongly agree}). We then calculated the Pearson correlations between GPT-4o-predicted and prompted agreement scores for each pair of example and estimated statements.

\subsubsection{Similarity calculation}

To examine the relationship between model response correlations and statement semantic similarities, we first collected the vector embeddings for all 64 statements using the OpenAI API, \texttt{text-embedding-3-large} embedding model. We then calculated the cosine similarities between all statement pairs' (n = 4,032) vector embeddings.

\[
S_C(\mathbf{A}, \mathbf{B}) = \frac{\sum_{i=1}^n A_i B_i}{\sqrt{\sum_{i=1}^n A_i^2} \cdot \sqrt{\sum_{i=1}^n B_i^2}}
\]

\subsection{Results}

\begin{figure}
    \centering
    \includegraphics[width=.8\linewidth]{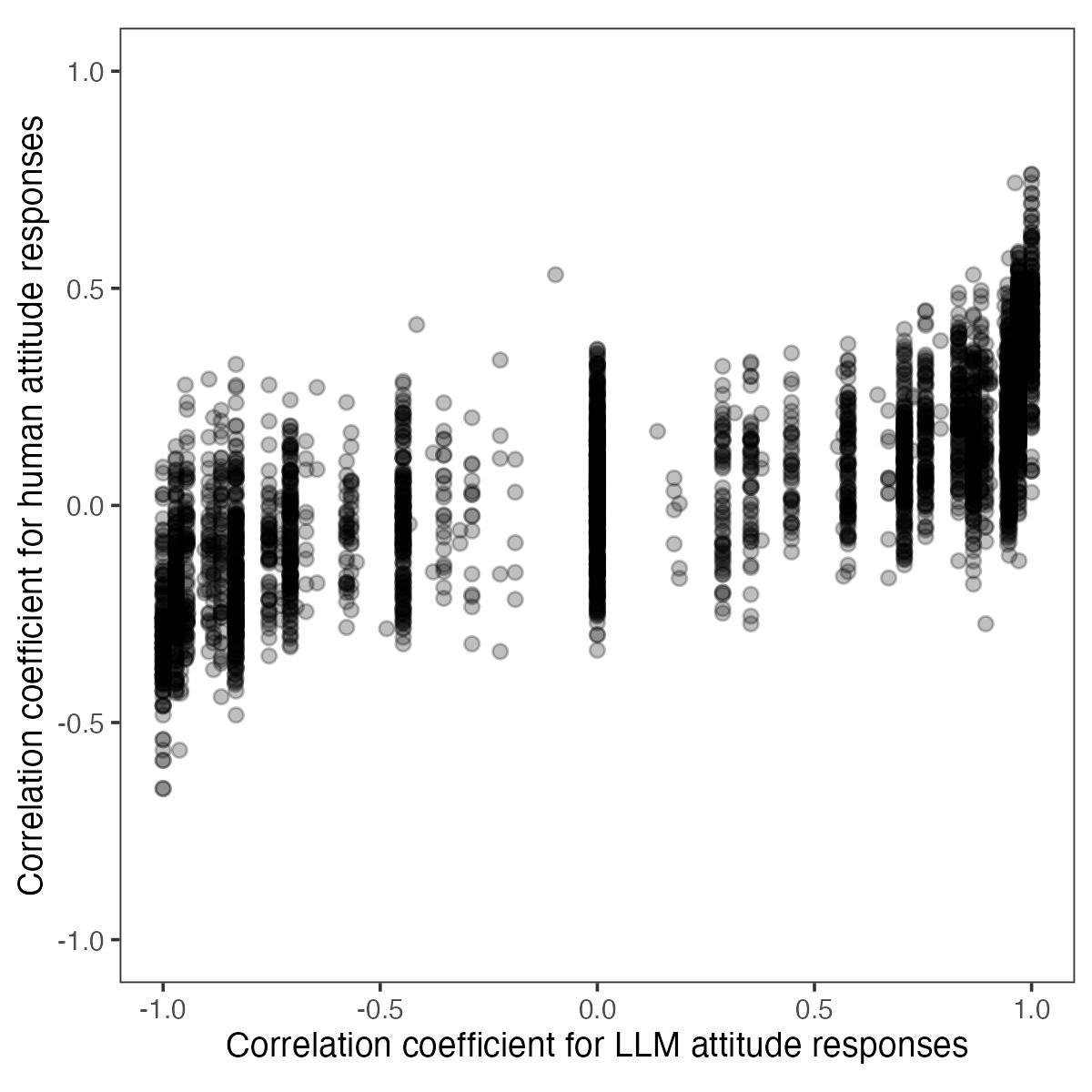}
    \caption{Scatterplot of correlation values observed in human data against correlation values among GPT-4o estimations.}
    \label{fig:human-llm scatterplot}
\end{figure}

\subsubsection{Human and GPT-4o-estimated correlations}

As shown in Figure \ref{fig:human-llm scatterplot}, GPT-4o was able to estimate human attitudes by capturing the inter-correlations among those attitudes. Comparing the observed correlations among human responses and the correlations estimated from GPT-4o's predictions, we found these coefficients to be themselves strongly and positively correlated with one another ($r = .77$, 95\% CI $[.76, .78]$, $p < .001$). However, whereas human attitude correlations were approximately normally distributed from -0.65 to 0.76, correlations estimated from GPT-4o tended to be more extreme, following a bimodal distribution with peaks near -1 and 1. Figure \ref{fig:human-llm scatterplot} illustrates this pattern in the clustering of points at the extremes along the x-axis. 
\\ % hack -- idk why we're needing this but it works
\subsubsection{Semantic Similarity and Estimated Correlations}

Next, we examined the extent to which GPT-4o's predicted associations between attitudes rely on the semantic similarity between those attitudes. Cosine similarity of the statements was positively correlated with the absolute strength of association among statements ($r = .33$, 95\% CI $[.30, .75]$, $p < .001$). This relationship is quite imperfect, so that dissimilar statements sometimes have strong associations, and similar statements sometimes have only weak associations. 

% \begin{figure}
%     \centering
%     \includegraphics[width=1\linewidth]{figures/cor-x-cosine.png}
%     \caption{Scatterplot of absolute correlation values observed in human data against the cosine distance of each statement pair.}
%     \label{fig:cor-cosine scatterplot}
% \end{figure}

\begin{figure}
    \centering
    \includegraphics[width=.8\linewidth]{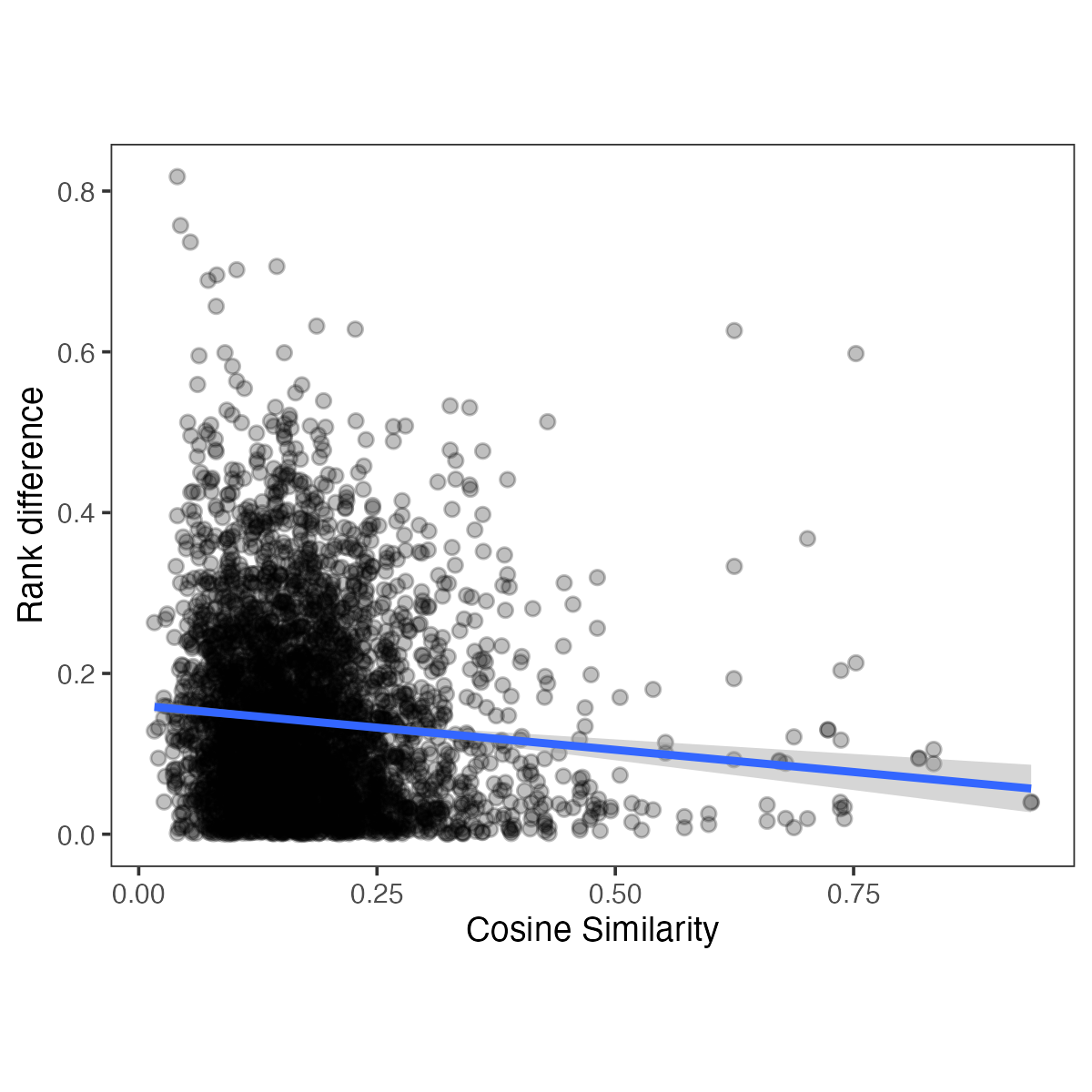}
    \caption{Scatterplot comparing differences between rank-transformed correlation values observed in human data and estimated by GPT-4o against the cosine distance of each statement pair. A regression line highlights the negative linear relationship.}
    \label{fig:rank-cosine scatterplot}
\end{figure}

As GPT-4o reliably overestimates the strength of associations, in the following analyses we rank-transform both human and LLM-predicted correlations for comparison purposes. Figure \ref{fig:rank-cosine scatterplot} shows the relationship between the rank-differences comparing human and LLM-estimated attitude correlation against their statement's cosine similarities. A modest negative trend is observed, indicating that model-estimated correlations are more reflective of human correlations among more-similar item pairs. 

A tendency to leverage similarity appears to bias GPT-4o's correlation estimates more generally: Regressing the absolute human correlation values on cosine similarity for each pair, we divide attitude-pairs into those more and less-strongly associated than would be predicted by cosine similarity. We find that GPT-4o more-commonly overestimates the association for pairs that are less-strongly associated than predicted from cosine similarity (64.9\% llm-based-rank $>$ human-rank) and underestimates associations for pairs that are more-strongly associated (70.5\% llm-based-rank $<$ human-rank). 

Nevertheless, the biases induced by semantic similarity appear to have only modest impacts on GPT-4o's ability to recapitulate correlations among attitudes. Filtering for only dissimilar (i.e. $S_c < .20$) attitude pairs significantly reduces the correlation between GPT-4o-estimated correlations and human correlations ($p = .001$), yet the relationship between estimated and observed correlations remained strong ($r = .724$). 

\section{Study 3: Predicting human responses using GPT-4o}

Finally, we examined GPT-4o's ability to predict individual respondents' answers to each target attitude from their responses to other attitude questions. Crucially, we are interested in examining the degree to which GPT-4o might rely upon the semantic similarity of the target attitude and the attitudes in its prompt.

\subsection{Selecting and evaluating top-k predictors}

To test this, we examined GPT-4o's ability to predict target attitudes using two feature selection methods. First, following \citeA{hwangAligningLanguageModels2023}, we identified a set of \textit{semantically-similar} predictors for each target attitude by selecting the top-k most similar attitudes based on cosine similarity (with $k \in \{3, 8\}$). For each target, we also identified a set of \textit{semantically-dissimilar} predictors. Here, we exclude any predictors with $S_c > .20$ and from the remainder choose the top-$k$ most correlated attitudes based on human data. 

\subsection{``Oracle'' model training}

For each target attitude and set of semantically-similar and dissimilar predictors, we trained a random forest model to examine how well those predictors could in-fact predict the target item. We evaluate the model fit using a 10-fold cross validation procedure, evaluating predictive accuracy using the model predictions on the held-out validation splits. We take these models as ``oracle'' models and we take their validation performance to represent the upper-bound of predictability for the target attitudes from the selected predictors.\footnote{We qualify this label with a potential limitation: given that these models are estimated from a limited sample of human data, superior predictive accuracy may technically be possible.}

\subsection{Generating GPT-4o predictions}

To predict each participant's responses, we constructed prompts containing 1) instructions for the model to "Help predict a person's answers on a social attitudes survey"; 2) a set of k (i.e., 3 or 8) example statements and a participant's actual responses observed in Study 1; 3) instructions to generate a prediction based on those statement-answer pairs; and 4) the target statement along with the list of answer choices (i.e., the five agreement levels). For each item, we composed predictor sets for the top-3 and top-8 items chosen using our criteria for selecting semantically-similar and semantically-dissimilar items. 
We then passed the prompts to GPT-4o through the OpenAI API.

\subsection{Metrics}

\subsubsection{Accuracy} For GPT-4o, we define accuracy as the exact match between the model's output and the choice text. 

\subsection{Results}

% \begin{table}
% \centering
% \caption{Predictive accuracies of GPT-4o and ``Oracle'' Random Forest models}

% \resizebox{\columnwidth}{!}{
% \begin{tabular}{ccccc}
% \toprule
% Model & Target subset & Selection method & Top-$k$ & Accuracy\\
% \midrule
% GPT-4o & all & similar ($S_c$) & 3 & 43.5\% \\
% GPT-4o & all &   & 8 & 45.2\% \\
% GPT-4o & all & dissimilar ($r$) & 3 & 40.4\% \\
% GPT-4o & all &  & 8 & 41.9\% \\
% \midrule
% GPT-4o & dissimilar $>$ similar & similar ($S_c$) & 3 & 41.2\% \\
% GPT-4o & dissimilar $>$ similar & similar ($S_c$) & 8 & 40.8\% \\
% GPT-4o & dissimilar $>$ similar & dissimilar ($r$) & 3 & 42.0\% \\
% GPT-4o & dissimilar $>$ similar & dissimilar ($r$) & 8 & 42.2\% \\
% \midrule
% Oracle & all & similar ($S_c$) & 3 & 47.7\% \\
% Oracle & all & similar ($S_c$) & 8 & 49.6\% \\
% Oracle & all & dissimilar ($r$) & 3 & 44.4\% \\
% Oracle & all & dissimilar ($r$) & 8 & 45.5\% \\
% \midrule
% Chance & all & -- & -- & 29.4\% \\
% \bottomrule
% \end{tabular}
% }
% \label{tab:prediction accuracy}
% \end{table}

\begin{table}
\caption{Predictive accuracies of GPT-4o and ``Oracle'' Random Forest models.\\}
\resizebox{\columnwidth}{!}{
\begin{tabular}{ccccc}
\toprule
Model & Target subset & Selection method & Top-$k$ & Accuracy\\
\midrule
GPT4o & all & similar ($S_c$) & 3 & 43.5\% \\
 &  &  & 8 & 45.2\% \\
 &  & dissimilar ($r$) & 3 & 40.4\% \\
 &  &  & 8 & 41.9\% \\
\midrule
GPT4o & dissimilar $>$ similar & similar ($S_c$) & 3 & 41.2\% \\
 &  &  & 8 & 40.8\% \\
 &  & dissimilar ($r$) & 3 & 42.0\% \\
 &  &  & 8 & 42.2\% \\
\midrule
Oracle & all & similar ($S_c$) & 3 & 47.7\% \\
 &  &  & 8 & 49.6\% \\
 &  & dissimilar ($r$) & 3 & 44.4\% \\
 &  &  & 8 & 45.5\% \\
 \midrule
Oracle & dissimilar $>$ similar & similar ($S_c$) & 3 & 43.9\% \\
 &  &  & 8 & 45.1\% \\
 &  & dissimilar ($r$) & 3 & 47.3\% \\
 &  &  & 8 & 47.6\% \\
\midrule
Chance & all & -- & -- & 29.4\% \\
\bottomrule
\end{tabular}

}
\label{tab:prediction accuracy}
\\ \textit{Note:} Models are evaluated against all observed responses from all participants ($64 \ \text{items} \times 376 \ \text{participants} = 24,064$).
\end{table}

Table \ref{tab:prediction accuracy} shows the predictive accuracy of "oracle" models and GPT-4o based on the top-k semantically-similar and semantically-dissimilar item selection strategies. As shown by the oracle model results, the semantically-similar items were in-fact more predictive of the target attitudes. This is to be expected: semantically-similar items often have relatively strong correlations with the target, so that excluding these items puts the dissimilar models at a substantial disadvantage, despite choosing the dissimilar items based on observed correlations. When ncluding all items, GPT-4o's predictions were more accurate when prompted with similar items. 
% GPT-4o's predictions were clearly more accurate than chance, but fell short of the oracle models.

Although the semantically-similar items were more predictive overall, for a substantial portion of target attitudes ($24/64$ for $k=3$ and $14/64$ for $k=8$), oracle models using semantically-dissimilar items were superior to those using similar items. Again this is generally to be expected: the most semantically-similar items are not \textit{always} those that are most predictive. Importantly, these cases offer strong a test of the role of semantic similarity in informing language models' attitude predictions: If the language model predicts as well or better with low-similarity items, this would then demonstrate that it is not relying on simple semantic similarity. However, if its predictions are worse in these cases where they could or should be better, then we can conclude it is in some way relying on semantic similarity.

Focusing on just these items, the results in Table \ref{tab:prediction accuracy} demonstrate that GPT-4o is also more accurate when predicting from the dissimilar items than similar items.  We take this as a clear demonstration that GPT-4o is capable of making meaningful social inferences in the absence of semantic similarity between statements. At the same time, comparing the accuracy of these two predictor sets between GPT-4o and the oracle models fit to human data reveals that GPT-4o's predictions are closer to the upper-bound of predictability when it is prompted with semantically-similar items (by 2.6\% for $k=3$ and 1.1\% for $k=8$). This suggests that GPT-4o does, at the same time, leverage the more surface-level similarity between attitudes in generating predictions. 
% An interesting question for future research would be to examine the potential for different computational pathways by which language models might make use of these different types of information.

% \begin{table}
%     \centering
%     \caption{Top-K GPT-4o prediction accuracies}
%     \begin{tabular}{lcc}
%     \toprule
%         Top-K metric & Top-K & Accuracy \\
%         \midrule
%         Cosine similarity & 3 & $43.52\%$ \\
%         Cosine similarity & 8 & $45.15\%$ \\
%         Human corr. & 3 & $40.37\%$ \\
%         Human corr. & 8 & $41.93\%$ \\
%         \bottomrule
%     \end{tabular}
%     \label{tab:prediction accuracy}
% \end{table}

\section{Discussion}

Using a novel dataset of human attitude across a range of social topics, we tested the ability of frontier language models (i.e. GPT-4o) to estimate the correlations among human attitudes and to predict individuals' attitudes from one another. Our findings demonstrate that GPT-4o can largely replicate the inter-correlations observed in human data, and can meaningfully predict individuals' attitudes. Going beyond prior findings \cite{hwangAligningLanguageModels2023}, we show that LLMs attitude predictions are not merely driven by surface-level semantic similarity, but instead tap in to latent associations among human attitudes.

Despite these novel findings, our study has a limitation that we only calculated model estimation alignment and prediction accuracy based on five levels of outputs for each target-example statement pair. Although we experimented with repeated sampling and prompting for probability distribution outputs, these approaches did not result in much variability or improvement in model estimations. 

Across our analyses with GPT-4o, we constrained the model to immediately respond with a predicted response option. This may underestimate the potential performance of the model, and future research could explore other prompting strategies. For instance, model prompts could utilize the chain-of-thought methods (CoT) \cite{wei2023chain} to encourage step-by-step ``thinking'' from the model, which might improve performance \cite{LongEtAl2024Aligning}. Alternately, future work might examine the capabilities of new reasoning models such as GPT-o1 \cite{openai2024} and DeepThink R1 \cite{deepseekai2025deepseekr1}. These approaches might improve performance or decrease the biases we observed with respect to semantic similarity. Further, examination of the intermediate CoT or reasoning steps might support deeper insights into the information leveraged by the model. For instance, we might examine whether models perform better when they explicitly infer (and tokenize in their reasoning) the ideology of a target person, or when they take other latent dimensions of attitudes into account.
% measurement of the relative contributions from surface-level semantic similarity and disentangle the predictive effect of models' latent, abstract representations. 

% Two, we are interested to work  with reasoning models such as GPT-o1 \cite{openai2024} and DeepThink R1 \cite{DeepThinkR1} to collect model predictions, so that we can document the underlying reasoning process additional to the prediction outcomes. By looking directly at the information at use for model predictions, we might compare LLMs’ approach with those in human social cognition. For instance, while humans can form certain attitudes based on self-interest \cite{Sears1991} or their personality \cite{Adorno1950}, we can observe if LLMs can make use of information in these specific ways. 

% Alternatively, we could explore methods other than model prompting. For instance, the percentage of Claude 3 Sonnet's anti-immigration selection increased continuously when a ``pro-life and anti-abortion stance'' feature was enhanced through a feature-steering method \cite{anthropic2024}. 

% underscore the existing call in the field for consideration on potential harmful practices LLM over-alignment could bring. 
Our findings underscore a number of existing concerns about the safe deployment of LLMs and other AI systems.
First, LLM alignment efforts seeking to personalize models to users' existing attitudes risk creating echo chambers in human-AI interactions, with the potential to increase polarization \cite{axelrod2021preventing}.
In addition, the major objective of monetizing LLM chatbots has driven large tech companies to adopt models for advertisement creation and promotion, e.g. Microsoft's Copilot \cite{microsoft2024privacy}. When AI ads are difficult to notice even with ad disclosures, users reported feeling manipulated, intruded, and less trusting towards LLM chatbots \cite{tang2024genai}. 

Moreover, AI safety advocates and researchers have identified a number of capabilities that, if achieved by AI systems, could present substantial risk to humans. Among these are the ability to persuade or manipulate humans \cite{JiEtAl2024AI, BurtellWoodside2023Artificial}. Persuasion often benefits from understanding of others' existing viewpoints \cite{lewandowskyMisinformationItsCorrection2012, PowellEtAl2023Modeling}, suggesting that the ability to predict attitudes may be a key aspect of this larger capacity. 
Our findings therefore underscore existing concerns about the potential risks of AI systems for persuasion and manipulation.
Although there may be positive uses of such capabilities \cite{karinshak2023working}, these capabilities nevertheless present alarming risks for automated propaganda and targeted misinformation.

\subsection{Conclusion}

Using a novel dataset of human responses to diverse attitude measures, we demonstrated the ability of frontier language models to replicate the interrelations among human beliefs without relying on surface semantic similarities. These results showcase these models' higher-level social reasoning capacity that extends beyond mere pattern matching. 

%% -------

% To test if GPT-4o is capable of making meaningful predictions without basing entirely on statement similarity, we collected model predictions based on either semantically similar or semantically dissimilar (but predictive of human correlations) predictors. Although model predictions were more accurate when provided with semantically similar items, it still produced meaningful predictions based on a subset of related attitudes.

% Previously, our understanding of LLMs’ ability to generalize user opinions was largely based on the surface-level similarity among closely-related attitudes \cite{hwangAligningLanguageModels2023, santurkarWhoseOpinionsLanguage2023}. Our work broadened this picture by demonstrating that GPT-4o, while partially relying on statement similarities, can represent some latent clusters of human attitudes and beliefs. 

% \section{Acknowledgments}

% In the \textbf{initial submission}, please \textbf{do not include
%   acknowledgements}, to preserve anonymity.  In the \textbf{final submission},
% place acknowledgments (including funding information) in a section \textbf{at
% the end of the paper}.

\bibliographystyle{apacite}

\setlength{\bibleftmargin}{.125in}
\setlength{\bibindent}{-\bibleftmargin}

\bibliography{references}

\begin{thebibliography}{}

\bibitem [\protect \citeauthoryear {%
Adorno%
, Frenkel-Brunswik%
, Levinson%
\BCBL {}\ \BBA {} Sanford%
}{%
Adorno%
\ \protect \BOthers {.}}{%
{\protect \APACyear {1950}}%
}]{%
Adorno1950}
\APACinsertmetastar {%
Adorno1950}%
\begin{APACrefauthors}%
Adorno, T\BPBI W.%
, Frenkel-Brunswik, E.%
, Levinson, D\BPBI J.%
\BCBL {}\ \BBA {} Sanford, N\BPBI R.%
\end{APACrefauthors}%
\unskip\
\newblock
\APACrefYear{1950}.
\newblock
\APACrefbtitle {The Authoritarian Personality} {The authoritarian personality}.
\newblock
\APACaddressPublisher{New York, NY}{Harpers}.
\PrintBackRefs{\CurrentBib}

\bibitem [\protect \citeauthoryear {%
Altemeyer%
}{%
Altemeyer%
}{%
{\protect \APACyear {1998}}%
}]{%
Altemeyer1998}
\APACinsertmetastar {%
Altemeyer1998}%
\begin{APACrefauthors}%
Altemeyer, B.%
\end{APACrefauthors}%
\unskip\
\newblock
\APACrefYearMonthDay{1998}{}{}.
\newblock
{\BBOQ}\APACrefatitle {The Other “Authoritarian Personality”} {The other “authoritarian personality”}.{\BBCQ}
\newblock
\BIn{} M\BPBI P.~Zanna\ (\BED), \APACrefbtitle {Advances in Experimental Social Psychology} {Advances in experimental social psychology}\ (\BVOL~30, \BPGS\ 47--92).
\newblock
\APACaddressPublisher{}{Academic Press}.
\newblock
\begin{APACrefDOI} \doi{10.1016/S0065-2601(08)60382-2} \end{APACrefDOI}
\PrintBackRefs{\CurrentBib}

\bibitem [\protect \citeauthoryear {%
Argyle%
\ \protect \BOthers {.}}{%
Argyle%
\ \protect \BOthers {.}}{%
{\protect \APACyear {2023}}%
}]{%
argyleOutOneMany2023}
\APACinsertmetastar {%
argyleOutOneMany2023}%
\begin{APACrefauthors}%
Argyle, L\BPBI P.%
, Busby, E\BPBI C.%
, Fulda, N.%
, Gubler, J\BPBI R.%
, Rytting, C.%
\BCBL {}\ \BBA {} Wingate, D.%
\end{APACrefauthors}%
\unskip\
\newblock
\APACrefYearMonthDay{2023}{{\APACmonth{02}}}{}.
\newblock
{\BBOQ}\APACrefatitle {Out of {{One}}, {{Many}}: {{Using Language Models}} to {{Simulate Human Samples}}} {Out of {{One}}, {{Many}}: {{Using Language Models}} to {{Simulate Human Samples}}}.{\BBCQ}
\newblock
\APACjournalVolNumPages{Political Analysis}{}{}{1--15}.
\newblock
\begin{APACrefDOI} \doi{10.1017/pan.2023.2} \end{APACrefDOI}
\PrintBackRefs{\CurrentBib}

\bibitem [\protect \citeauthoryear {%
Axelrod%
, Daymude%
\BCBL {}\ \BBA {} Forrest%
}{%
Axelrod%
\ \protect \BOthers {.}}{%
{\protect \APACyear {2021}}%
}]{%
axelrod2021preventing}
\APACinsertmetastar {%
axelrod2021preventing}%
\begin{APACrefauthors}%
Axelrod, R.%
, Daymude, J\BPBI J.%
\BCBL {}\ \BBA {} Forrest, S.%
\end{APACrefauthors}%
\unskip\
\newblock
\APACrefYearMonthDay{2021}{}{}.
\newblock
{\BBOQ}\APACrefatitle {Preventing extreme polarization of political attitudes} {Preventing extreme polarization of political attitudes}.{\BBCQ}
\newblock
\APACjournalVolNumPages{Proceedings of the National Academy of Sciences}{118}{50}{e2102139118}.
\newblock
\begin{APACrefURL} \url{https://doi.org/10.1073/pnas.2102139118} \end{APACrefURL}
\newblock
\begin{APACrefDOI} \doi{10.1073/pnas.2102139118} \end{APACrefDOI}
\PrintBackRefs{\CurrentBib}

\bibitem [\protect \citeauthoryear {%
Burtell%
\ \BBA {} Woodside%
}{%
Burtell%
\ \BBA {} Woodside%
}{%
{\protect \APACyear {2023}}%
}]{%
BurtellWoodside2023Artificial}
\APACinsertmetastar {%
BurtellWoodside2023Artificial}%
\begin{APACrefauthors}%
Burtell, M.%
\BCBT {}\ \BBA {} Woodside, T.%
\end{APACrefauthors}%
\unskip\
\newblock
\APACrefYearMonthDay{2023}{{\APACmonth{03}}}{}.
\newblock
\APACrefbtitle {Artificial {{Influence}}: {{An Analysis Of AI-Driven Persuasion}}} {Artificial {{Influence}}: {{An Analysis Of AI-Driven Persuasion}}}\ (\BNUM\ arXiv:2303.08721).
\newblock
\APACaddressPublisher{}{arXiv}.
\newblock
\begin{APACrefDOI} \doi{10.48550/arXiv.2303.08721} \end{APACrefDOI}
\PrintBackRefs{\CurrentBib}

\bibitem [\protect \citeauthoryear {%
Byrne%
}{%
Byrne%
}{%
{\protect \APACyear {1961}}%
}]{%
Byrne1961}
\APACinsertmetastar {%
Byrne1961}%
\begin{APACrefauthors}%
Byrne, D.%
\end{APACrefauthors}%
\unskip\
\newblock
\APACrefYearMonthDay{1961}{}{}.
\newblock
{\BBOQ}\APACrefatitle {Interpersonal Attraction and Attitude Similarity} {Interpersonal attraction and attitude similarity}.{\BBCQ}
\newblock
\APACjournalVolNumPages{The Journal of Abnormal and Social Psychology}{62}{3}{713--715}.
\newblock
\begin{APACrefDOI} \doi{10.1037/h0044721} \end{APACrefDOI}
\PrintBackRefs{\CurrentBib}

\bibitem [\protect \citeauthoryear {%
Choenni%
\ \BBA {} Shutova%
}{%
Choenni%
\ \BBA {} Shutova%
}{%
{\protect \APACyear {2024}}%
}]{%
ChoenniShutova2024SelfAlignment}
\APACinsertmetastar {%
ChoenniShutova2024SelfAlignment}%
\begin{APACrefauthors}%
Choenni, R.%
\BCBT {}\ \BBA {} Shutova, E.%
\end{APACrefauthors}%
\unskip\
\newblock
\APACrefYearMonthDay{2024}{{\APACmonth{08}}}{}.
\newblock
\APACrefbtitle {Self-{{Alignment}}: {{Improving Alignment}} of {{Cultural Values}} in {{LLMs}} via {{In-Context Learning}}} {Self-{{Alignment}}: {{Improving Alignment}} of {{Cultural Values}} in {{LLMs}} via {{In-Context Learning}}}\ (\BNUM\ arXiv:2408.16482).
\newblock
\APACaddressPublisher{}{arXiv}.
\newblock
\begin{APACrefDOI} \doi{10.48550/arXiv.2408.16482} \end{APACrefDOI}
\PrintBackRefs{\CurrentBib}

\bibitem [\protect \citeauthoryear {%
Converse%
}{%
Converse%
}{%
{\protect \APACyear {1964}}%
}]{%
Converse1964}
\APACinsertmetastar {%
Converse1964}%
\begin{APACrefauthors}%
Converse, P\BPBI E.%
\end{APACrefauthors}%
\unskip\
\newblock
\APACrefYearMonthDay{1964}{}{}.
\newblock
{\BBOQ}\APACrefatitle {The nature of belief systems in mass publics (1964)} {The nature of belief systems in mass publics (1964)}.{\BBCQ}
\newblock
\APACjournalVolNumPages{Critical Review}{18}{1–3}{1--74}.
\newblock
\begin{APACrefDOI} \doi{10.1080/08913810608443650} \end{APACrefDOI}
\PrintBackRefs{\CurrentBib}

\bibitem [\protect \citeauthoryear {%
{DeepSeek-AI}%
\ \protect \BOthers {.}}{%
{DeepSeek-AI}%
\ \protect \BOthers {.}}{%
{\protect \APACyear {2025}}%
}]{%
deepseekai2025deepseekr1}
\APACinsertmetastar {%
deepseekai2025deepseekr1}%
\begin{APACrefauthors}%
{DeepSeek-AI}%
, Guo, D.%
, Yang, D.%
, Zhang, H.%
, Song, J.%
, Zhang, R.%
\BDBL {}Zhang, Z.%
\end{APACrefauthors}%
\unskip\
\newblock
\APACrefYearMonthDay{2025}{}{}.
\newblock
\APACrefbtitle {DeepSeek-R1: Incentivizing Reasoning Capability in LLMs via Reinforcement Learning} {Deepseek-r1: Incentivizing reasoning capability in llms via reinforcement learning}\ (\BNUM\ arXiv:2501.12948).
\newblock
\begin{APACrefURL} \url{https://doi.org/10.48550/arXiv.2501.12948} \end{APACrefURL}
\newblock
\APACrefnote{arXiv preprint}
\newblock
\begin{APACrefDOI} \doi{10.48550/arXiv.2501.12948} \end{APACrefDOI}
\PrintBackRefs{\CurrentBib}

\bibitem [\protect \citeauthoryear {%
{El-Sayed}%
\ \protect \BOthers {.}}{%
{El-Sayed}%
\ \protect \BOthers {.}}{%
{\protect \APACyear {2024}}%
}]{%
El-SayedEtAl2024MechanismBased}
\APACinsertmetastar {%
El-SayedEtAl2024MechanismBased}%
\begin{APACrefauthors}%
{El-Sayed}, S.%
, Akbulut, C.%
, McCroskery, A.%
, Keeling, G.%
, Kenton, Z.%
, Jalan, Z.%
\BDBL {}Brown, S.%
\end{APACrefauthors}%
\unskip\
\newblock
\APACrefYearMonthDay{2024}{{\APACmonth{04}}}{}.
\newblock
\APACrefbtitle {A {{Mechanism-Based Approach}} to {{Mitigating Harms}} from {{Persuasive Generative AI}}} {A {{Mechanism-Based Approach}} to {{Mitigating Harms}} from {{Persuasive Generative AI}}}\ (\BNUM\ arXiv:2404.15058).
\newblock
\APACaddressPublisher{}{arXiv}.
\newblock
\begin{APACrefDOI} \doi{10.48550/arXiv.2404.15058} \end{APACrefDOI}
\PrintBackRefs{\CurrentBib}

\bibitem [\protect \citeauthoryear {%
Festinger%
}{%
Festinger%
}{%
{\protect \APACyear {1957}}%
}]{%
festinger1957cognitive}
\APACinsertmetastar {%
festinger1957cognitive}%
\begin{APACrefauthors}%
Festinger, L.%
\end{APACrefauthors}%
\unskip\
\newblock
\APACrefYear{1957}.
\newblock
\APACrefbtitle {{A Theory of Cognitive Dissonance}} {{A Theory of Cognitive Dissonance}}.
\newblock
\APACaddressPublisher{}{Stanford University Press}.
\newblock
\APAChowpublished {Paperback}.
\PrintBackRefs{\CurrentBib}

\bibitem [\protect \citeauthoryear {%
Gerstenberg%
\ \BBA {} Tenenbaum%
}{%
Gerstenberg%
\ \BBA {} Tenenbaum%
}{%
{\protect \APACyear {2017}}%
}]{%
gerstenbergIntuitiveTheories2017}
\APACinsertmetastar {%
gerstenbergIntuitiveTheories2017}%
\begin{APACrefauthors}%
Gerstenberg, T.%
\BCBT {}\ \BBA {} Tenenbaum, J\BPBI B.%
\end{APACrefauthors}%
\unskip\
\newblock
\APACrefYearMonthDay{2017}{}{}.
\newblock
{\BBOQ}\APACrefatitle {Intuitive Theories} {Intuitive theories}.{\BBCQ}
\newblock
\BIn{} \APACrefbtitle {The {{Oxford}} Handbook of Causal Reasoning} {The {{Oxford}} handbook of causal reasoning}\ (\BPGS\ 515--547).
\newblock
\APACaddressPublisher{New York, NY, US}{Oxford University Press}.
\newblock
\begin{APACrefDOI} \doi{10.1093/oxfordhb/9780199399550.001.0001} \end{APACrefDOI}
\PrintBackRefs{\CurrentBib}

\bibitem [\protect \citeauthoryear {%
Hofstede%
}{%
Hofstede%
}{%
{\protect \APACyear {2001}}%
}]{%
Hofstede2001}
\APACinsertmetastar {%
Hofstede2001}%
\begin{APACrefauthors}%
Hofstede, G.%
\end{APACrefauthors}%
\unskip\
\newblock
\APACrefYear{2001}.
\newblock
\APACrefbtitle {Culture’s Consequences: Comparing Values, Behaviors, Institutions, and Organizations Across Nations} {Culture’s consequences: Comparing values, behaviors, institutions, and organizations across nations}.
\newblock
\APACaddressPublisher{Thousand Oaks, CA}{Sage Publications}.
\newblock
\begin{APACrefDOI} \doi{10.1016/S0005-7967(02)00184-5} \end{APACrefDOI}
\PrintBackRefs{\CurrentBib}

\bibitem [\protect \citeauthoryear {%
Holyoak%
\ \BBA {} Simon%
}{%
Holyoak%
\ \BBA {} Simon%
}{%
{\protect \APACyear {1999}}%
}]{%
holyoakBidirectionalReasoningDecision1999}
\APACinsertmetastar {%
holyoakBidirectionalReasoningDecision1999}%
\begin{APACrefauthors}%
Holyoak, K\BPBI J.%
\BCBT {}\ \BBA {} Simon, D.%
\end{APACrefauthors}%
\unskip\
\newblock
\APACrefYearMonthDay{1999}{}{}.
\newblock
{\BBOQ}\APACrefatitle {Bidirectional {{Reasoning}} in {{Decision Making}} by {{Constraint Satisfaction}}} {Bidirectional {{Reasoning}} in {{Decision Making}} by {{Constraint Satisfaction}}}.{\BBCQ}
\newblock
\APACjournalVolNumPages{Journal of Experimental Psychology: General}{}{}{29}.
\PrintBackRefs{\CurrentBib}

\bibitem [\protect \citeauthoryear {%
Hwang%
, Majumder%
\BCBL {}\ \BBA {} Tandon%
}{%
Hwang%
\ \protect \BOthers {.}}{%
{\protect \APACyear {2023}}%
}]{%
hwangAligningLanguageModels2023}
\APACinsertmetastar {%
hwangAligningLanguageModels2023}%
\begin{APACrefauthors}%
Hwang, E.%
, Majumder, B\BPBI P.%
\BCBL {}\ \BBA {} Tandon, N.%
\end{APACrefauthors}%
\unskip\
\newblock
\APACrefYearMonthDay{2023}{{\APACmonth{05}}}{}.
\newblock
\APACrefbtitle {Aligning {{Language Models}} to {{User Opinions}}} {Aligning {{Language Models}} to {{User Opinions}}}\ (\BNUM\ arXiv:2305.14929).
\newblock
\APACaddressPublisher{}{arXiv}.
\PrintBackRefs{\CurrentBib}

\bibitem [\protect \citeauthoryear {%
Ji%
\ \protect \BOthers {.}}{%
Ji%
\ \protect \BOthers {.}}{%
{\protect \APACyear {2024}}%
}]{%
JiEtAl2024AI}
\APACinsertmetastar {%
JiEtAl2024AI}%
\begin{APACrefauthors}%
Ji, J.%
, Qiu, T.%
, Chen, B.%
, Zhang, B.%
, Lou, H.%
, Wang, K.%
\BDBL {}Gao, W.%
\end{APACrefauthors}%
\unskip\
\newblock
\APACrefYearMonthDay{2024}{{\APACmonth{05}}}{}.
\newblock
\APACrefbtitle {{{AI Alignment}}: {{A Comprehensive Survey}}} {{{AI Alignment}}: {{A Comprehensive Survey}}}\ (\BNUM\ arXiv:2310.19852).
\newblock
\APACaddressPublisher{}{arXiv}.
\newblock
\begin{APACrefDOI} \doi{10.48550/arXiv.2310.19852} \end{APACrefDOI}
\PrintBackRefs{\CurrentBib}

\bibitem [\protect \citeauthoryear {%
Jost%
, Glaser%
, Kruglanski%
\BCBL {}\ \BBA {} Sulloway%
}{%
Jost%
, Glaser%
\BCBL {}\ \protect \BOthers {.}}{%
{\protect \APACyear {2003}}%
}]{%
Jost2003Conservatism}
\APACinsertmetastar {%
Jost2003Conservatism}%
\begin{APACrefauthors}%
Jost, J\BPBI T.%
, Glaser, J.%
, Kruglanski, A\BPBI W.%
\BCBL {}\ \BBA {} Sulloway, F\BPBI J.%
\end{APACrefauthors}%
\unskip\
\newblock
\APACrefYearMonthDay{2003}{}{}.
\newblock
{\BBOQ}\APACrefatitle {Political Conservatism as Motivated Social Cognition} {Political conservatism as motivated social cognition}.{\BBCQ}
\newblock
\APACjournalVolNumPages{Psychological Bulletin}{129}{3}{339--375}.
\newblock
\begin{APACrefDOI} \doi{10.1037/0033-2909.129.3.339} \end{APACrefDOI}
\PrintBackRefs{\CurrentBib}

\bibitem [\protect \citeauthoryear {%
Jost%
, Pelham%
, Sheldon%
\BCBL {}\ \BBA {} Sullivan%
}{%
Jost%
, Pelham%
\BCBL {}\ \protect \BOthers {.}}{%
{\protect \APACyear {2003}}%
}]{%
Jost2003}
\APACinsertmetastar {%
Jost2003}%
\begin{APACrefauthors}%
Jost, J\BPBI T.%
, Pelham, B\BPBI W.%
, Sheldon, O.%
\BCBL {}\ \BBA {} Sullivan, B.%
\end{APACrefauthors}%
\unskip\
\newblock
\APACrefYearMonthDay{2003}{}{}.
\newblock
{\BBOQ}\APACrefatitle {Social inequality and the reduction of ideological dissonance on behalf of the system: Evidence of enhanced system justification among the disadvantaged} {Social inequality and the reduction of ideological dissonance on behalf of the system: Evidence of enhanced system justification among the disadvantaged}.{\BBCQ}
\newblock
\APACjournalVolNumPages{European Journal of Social Psychology}{33}{}{13--36}.
\newblock
\begin{APACrefDOI} \doi{10.1002/ejsp.127} \end{APACrefDOI}
\PrintBackRefs{\CurrentBib}

\bibitem [\protect \citeauthoryear {%
Karinshak%
, Liu%
, Park%
\BCBL {}\ \BBA {} Hancock%
}{%
Karinshak%
\ \protect \BOthers {.}}{%
{\protect \APACyear {2023}}%
}]{%
karinshak2023working}
\APACinsertmetastar {%
karinshak2023working}%
\begin{APACrefauthors}%
Karinshak, E.%
, Liu, S\BPBI X.%
, Park, J\BPBI S.%
\BCBL {}\ \BBA {} Hancock, J\BPBI T.%
\end{APACrefauthors}%
\unskip\
\newblock
\APACrefYearMonthDay{2023}{}{}.
\newblock
{\BBOQ}\APACrefatitle {Working With AI to Persuade: Examining a Large Language Model’s Ability to Generate Pro-Vaccination Messages} {Working with ai to persuade: Examining a large language model’s ability to generate pro-vaccination messages}.{\BBCQ}
\newblock
\APACjournalVolNumPages{Proceedings of the ACM on Human-Computer Interaction}{7}{CSCW1}{1--29}.
\newblock
\begin{APACrefDOI} \doi{10.1145/3579592} \end{APACrefDOI}
\PrintBackRefs{\CurrentBib}

\bibitem [\protect \citeauthoryear {%
Lewandowsky%
, Ecker%
, Seifert%
, Schwarz%
\BCBL {}\ \BBA {} Cook%
}{%
Lewandowsky%
\ \protect \BOthers {.}}{%
{\protect \APACyear {2012}}%
}]{%
lewandowskyMisinformationItsCorrection2012}
\APACinsertmetastar {%
lewandowskyMisinformationItsCorrection2012}%
\begin{APACrefauthors}%
Lewandowsky, S.%
, Ecker, U\BPBI K\BPBI H.%
, Seifert, C\BPBI M.%
, Schwarz, N.%
\BCBL {}\ \BBA {} Cook, J.%
\end{APACrefauthors}%
\unskip\
\newblock
\APACrefYearMonthDay{2012}{{\APACmonth{12}}}{}.
\newblock
{\BBOQ}\APACrefatitle {Misinformation and {{Its Correction}}: {{Continued Influence}} and {{Successful Debiasing}}} {Misinformation and {{Its Correction}}: {{Continued Influence}} and {{Successful Debiasing}}}.{\BBCQ}
\newblock
\APACjournalVolNumPages{Psychological Science in the Public Interest}{13}{3}{106--131}.
\newblock
\begin{APACrefDOI} \doi{10.1177/1529100612451018} \end{APACrefDOI}
\PrintBackRefs{\CurrentBib}

\bibitem [\protect \citeauthoryear {%
Long%
, Kawaguchi%
, Kan%
\BCBL {}\ \BBA {} Chen%
}{%
Long%
\ \protect \BOthers {.}}{%
{\protect \APACyear {2024}}%
}]{%
LongEtAl2024Aligning}
\APACinsertmetastar {%
LongEtAl2024Aligning}%
\begin{APACrefauthors}%
Long, D\BPBI X.%
, Kawaguchi, K.%
, Kan, M\BHBI Y.%
\BCBL {}\ \BBA {} Chen, N\BPBI F.%
\end{APACrefauthors}%
\unskip\
\newblock
\APACrefYearMonthDay{2024}{{\APACmonth{12}}}{}.
\newblock
\APACrefbtitle {Aligning {{Large Language Models}} with {{Human Opinions}} through {{Persona Selection}} and {{Value--Belief--Norm Reasoning}}} {Aligning {{Large Language Models}} with {{Human Opinions}} through {{Persona Selection}} and {{Value--Belief--Norm Reasoning}}}\ (\BNUM\ arXiv:2311.08385).
\newblock
\APACaddressPublisher{}{arXiv}.
\newblock
\begin{APACrefDOI} \doi{10.48550/arXiv.2311.08385} \end{APACrefDOI}
\PrintBackRefs{\CurrentBib}

\bibitem [\protect \citeauthoryear {%
McPherson%
, Smith-Lovin%
\BCBL {}\ \BBA {} Cook%
}{%
McPherson%
\ \protect \BOthers {.}}{%
{\protect \APACyear {2001}}%
}]{%
McPherson2001}
\APACinsertmetastar {%
McPherson2001}%
\begin{APACrefauthors}%
McPherson, M.%
, Smith-Lovin, L.%
\BCBL {}\ \BBA {} Cook, J\BPBI M.%
\end{APACrefauthors}%
\unskip\
\newblock
\APACrefYearMonthDay{2001}{}{}.
\newblock
{\BBOQ}\APACrefatitle {Birds of a Feather: Homophily in Social Networks} {Birds of a feather: Homophily in social networks}.{\BBCQ}
\newblock
\APACjournalVolNumPages{Annual Review of Sociology}{27}{}{415--444}.
\newblock
\begin{APACrefURL} \url{https://www.jstor.org/stable/2678628} \end{APACrefURL}
\newblock
\begin{APACrefDOI} \doi{10.1146/annurev.soc.27.1.415} \end{APACrefDOI}
\PrintBackRefs{\CurrentBib}

\bibitem [\protect \citeauthoryear {%
Microsoft%
}{%
Microsoft%
}{%
{\protect \APACyear {2024}}%
}]{%
microsoft2024privacy}
\APACinsertmetastar {%
microsoft2024privacy}%
\begin{APACrefauthors}%
Microsoft.%
\end{APACrefauthors}%
\unskip\
\newblock
\APACrefYearMonthDay{2024}{}{}.
\newblock
\APACrefbtitle {Microsoft Privacy Statement – Microsoft privacy.} {Microsoft privacy statement – microsoft privacy.}
\PrintBackRefs{\CurrentBib}

\bibitem [\protect \citeauthoryear {%
Moussaïd%
, Kämmer%
, Analytis%
\BCBL {}\ \BBA {} Neth%
}{%
Moussaïd%
\ \protect \BOthers {.}}{%
{\protect \APACyear {2013}}%
}]{%
Moussaid2013Social}
\APACinsertmetastar {%
Moussaid2013Social}%
\begin{APACrefauthors}%
Moussaïd, M.%
, Kämmer, J\BPBI E.%
, Analytis, P\BPBI P.%
\BCBL {}\ \BBA {} Neth, H.%
\end{APACrefauthors}%
\unskip\
\newblock
\APACrefYearMonthDay{2013}{}{}.
\newblock
{\BBOQ}\APACrefatitle {Social influence and the collective dynamics of opinion formation} {Social influence and the collective dynamics of opinion formation}.{\BBCQ}
\newblock
\APACjournalVolNumPages{PLOS ONE}{8}{11}{e78433}.
\newblock
\begin{APACrefDOI} \doi{10.1371/journal.pone.0078433} \end{APACrefDOI}
\PrintBackRefs{\CurrentBib}

\bibitem [\protect \citeauthoryear {%
OpenAI%
}{%
OpenAI%
}{%
{\protect \APACyear {2024}}%
}]{%
openai2024}
\APACinsertmetastar {%
openai2024}%
\begin{APACrefauthors}%
OpenAI.%
\end{APACrefauthors}%
\unskip\
\newblock
\APACrefYearMonthDay{2024}{}{}.
\newblock
\APACrefbtitle {Introducing OpenAI o1.} {Introducing openai o1.}
\newblock
\begin{APACrefURL} \url{https://openai.com/o1} \end{APACrefURL}
\PrintBackRefs{\CurrentBib}

\bibitem [\protect \citeauthoryear {%
OpenAI%
\ \protect \BOthers {.}}{%
OpenAI%
\ \protect \BOthers {.}}{%
{\protect \APACyear {2024}}%
}]{%
OpenAIEtAl2024GPT4o}
\APACinsertmetastar {%
OpenAIEtAl2024GPT4o}%
\begin{APACrefauthors}%
OpenAI%
, Hurst, A.%
, Lerer, A.%
, Goucher, A\BPBI P.%
, Perelman, A.%
, Ramesh, A.%
\BDBL {}Malkov, Y.%
\end{APACrefauthors}%
\unskip\
\newblock
\APACrefYearMonthDay{2024}{{\APACmonth{10}}}{}.
\newblock
\APACrefbtitle {{{GPT-4o System Card}}} {{{GPT-4o System Card}}}\ (\BNUM\ arXiv:2410.21276).
\newblock
\APACaddressPublisher{}{arXiv}.
\newblock
\begin{APACrefDOI} \doi{10.48550/arXiv.2410.21276} \end{APACrefDOI}
\PrintBackRefs{\CurrentBib}

\bibitem [\protect \citeauthoryear {%
Pawar%
\ \protect \BOthers {.}}{%
Pawar%
\ \protect \BOthers {.}}{%
{\protect \APACyear {2024}}%
}]{%
PawarEtAl2024Survey}
\APACinsertmetastar {%
PawarEtAl2024Survey}%
\begin{APACrefauthors}%
Pawar, S.%
, Park, J.%
, Jin, J.%
, Arora, A.%
, Myung, J.%
, Yadav, S.%
\BDBL {}Augenstein, I.%
\end{APACrefauthors}%
\unskip\
\newblock
\APACrefYearMonthDay{2024}{{\APACmonth{10}}}{}.
\newblock
\APACrefbtitle {Survey of {{Cultural Awareness}} in {{Language Models}}: {{Text}} and {{Beyond}}} {Survey of {{Cultural Awareness}} in {{Language Models}}: {{Text}} and {{Beyond}}}\ (\BNUM\ arXiv:2411.00860).
\newblock
\APACaddressPublisher{}{arXiv}.
\newblock
\begin{APACrefDOI} \doi{10.48550/arXiv.2411.00860} \end{APACrefDOI}
\PrintBackRefs{\CurrentBib}

\bibitem [\protect \citeauthoryear {%
Powell%
, Weisman%
\BCBL {}\ \BBA {} Markman%
}{%
Powell%
\ \protect \BOthers {.}}{%
{\protect \APACyear {2023}}%
}]{%
PowellEtAl2023Modeling}
\APACinsertmetastar {%
PowellEtAl2023Modeling}%
\begin{APACrefauthors}%
Powell, D.%
, Weisman, K.%
\BCBL {}\ \BBA {} Markman, E\BPBI M.%
\end{APACrefauthors}%
\unskip\
\newblock
\APACrefYearMonthDay{2023}{{\APACmonth{05}}}{}.
\newblock
{\BBOQ}\APACrefatitle {Modeling and Leveraging Intuitive Theories to Improve Vaccine Attitudes} {Modeling and leveraging intuitive theories to improve vaccine attitudes}.{\BBCQ}
\newblock
\APACjournalVolNumPages{Journal of Experimental Psychology: General}{152}{5}{1379--1395}.
\newblock
\begin{APACrefDOI} \doi{10.1037/xge0001324} \end{APACrefDOI}
\PrintBackRefs{\CurrentBib}

\bibitem [\protect \citeauthoryear {%
Pratto%
}{%
Pratto%
}{%
{\protect \APACyear {1999}}%
}]{%
Pratto1999}
\APACinsertmetastar {%
Pratto1999}%
\begin{APACrefauthors}%
Pratto, F.%
\end{APACrefauthors}%
\unskip\
\newblock
\APACrefYearMonthDay{1999}{}{}.
\newblock
{\BBOQ}\APACrefatitle {The Puzzle of Continuing Group Inequality: Piecing Together Psychological, Social, and Cultural Forces in Social Dominance Theory} {The puzzle of continuing group inequality: Piecing together psychological, social, and cultural forces in social dominance theory}.{\BBCQ}
\newblock
\APACjournalVolNumPages{Advances in Experimental Social Psychology}{31}{}{191--263}.
\PrintBackRefs{\CurrentBib}

\bibitem [\protect \citeauthoryear {%
Razafinirina%
, Dimbisoa%
\BCBL {}\ \BBA {} Mahatody%
}{%
Razafinirina%
\ \protect \BOthers {.}}{%
{\protect \APACyear {2024}}%
}]{%
RazafinirinaEtAl2024Pedagogical}
\APACinsertmetastar {%
RazafinirinaEtAl2024Pedagogical}%
\begin{APACrefauthors}%
Razafinirina, M\BPBI A.%
, Dimbisoa, W\BPBI G.%
\BCBL {}\ \BBA {} Mahatody, T.%
\end{APACrefauthors}%
\unskip\
\newblock
\APACrefYearMonthDay{2024}{}{}.
\newblock
{\BBOQ}\APACrefatitle {Pedagogical {{Alignment}} of {{Large Language Models}} ({{LLM}}) for {{Personalized Learning}}: {{A Survey}}, {{Trends}} and {{Challenges}}} {Pedagogical {{Alignment}} of {{Large Language Models}} ({{LLM}}) for {{Personalized Learning}}: {{A Survey}}, {{Trends}} and {{Challenges}}}.{\BBCQ}
\newblock
\APACjournalVolNumPages{Journal of Intelligent Learning Systems and Applications}{16}{04}{448--480}.
\newblock
\begin{APACrefDOI} \doi{10.4236/jilsa.2024.164023} \end{APACrefDOI}
\PrintBackRefs{\CurrentBib}

\bibitem [\protect \citeauthoryear {%
Santurkar%
\ \protect \BOthers {.}}{%
Santurkar%
\ \protect \BOthers {.}}{%
{\protect \APACyear {2023}}%
}]{%
santurkarWhoseOpinionsLanguage2023}
\APACinsertmetastar {%
santurkarWhoseOpinionsLanguage2023}%
\begin{APACrefauthors}%
Santurkar, S.%
, Durmus, E.%
, Ladhak, F.%
, Lee, C.%
, Liang, P.%
\BCBL {}\ \BBA {} Hashimoto, T.%
\end{APACrefauthors}%
\unskip\
\newblock
\APACrefYearMonthDay{2023}{{\APACmonth{03}}}{}.
\newblock
\APACrefbtitle {Whose {{Opinions Do Language Models Reflect}}?} {Whose {{Opinions Do Language Models Reflect}}?}\ (\BNUM\ arXiv:2303.17548).
\newblock
\APACaddressPublisher{}{arXiv}.
\PrintBackRefs{\CurrentBib}

\bibitem [\protect \citeauthoryear {%
Schachter%
}{%
Schachter%
}{%
{\protect \APACyear {1959}}%
}]{%
Schachter1959}
\APACinsertmetastar {%
Schachter1959}%
\begin{APACrefauthors}%
Schachter, S.%
\end{APACrefauthors}%
\unskip\
\newblock
\APACrefYear{1959}.
\newblock
\APACrefbtitle {The Psychology of Affiliation} {The psychology of affiliation}.
\newblock
\APACaddressPublisher{Stanford, CA}{Stanford University Press}.
\PrintBackRefs{\CurrentBib}

\bibitem [\protect \citeauthoryear {%
Sonkar%
, Ni%
, Chaudhary%
\BCBL {}\ \BBA {} Baraniuk%
}{%
Sonkar%
\ \protect \BOthers {.}}{%
{\protect \APACyear {2024}}%
}]{%
SonkarEtAl2024Pedagogical}
\APACinsertmetastar {%
SonkarEtAl2024Pedagogical}%
\begin{APACrefauthors}%
Sonkar, S.%
, Ni, K.%
, Chaudhary, S.%
\BCBL {}\ \BBA {} Baraniuk, R\BPBI G.%
\end{APACrefauthors}%
\unskip\
\newblock
\APACrefYearMonthDay{2024}{{\APACmonth{10}}}{}.
\newblock
\APACrefbtitle {Pedagogical {{Alignment}} of {{Large Language Models}}} {Pedagogical {{Alignment}} of {{Large Language Models}}}\ (\BNUM\ arXiv:2402.05000).
\newblock
\APACaddressPublisher{}{arXiv}.
\newblock
\begin{APACrefDOI} \doi{10.48550/arXiv.2402.05000} \end{APACrefDOI}
\PrintBackRefs{\CurrentBib}

\bibitem [\protect \citeauthoryear {%
Spellman%
, Ullman%
\BCBL {}\ \BBA {} Holyoak%
}{%
Spellman%
\ \protect \BOthers {.}}{%
{\protect \APACyear {1993}}%
}]{%
spellmanCoherenceModelCognitive1993}
\APACinsertmetastar {%
spellmanCoherenceModelCognitive1993}%
\begin{APACrefauthors}%
Spellman, B\BPBI A.%
, Ullman, J\BPBI B.%
\BCBL {}\ \BBA {} Holyoak, K\BPBI J.%
\end{APACrefauthors}%
\unskip\
\newblock
\APACrefYearMonthDay{1993}{{\APACmonth{01}}}{}.
\newblock
{\BBOQ}\APACrefatitle {A {{Coherence Model}} of {{Cognitive Consistency}}: {{Dynamics}} of {{Attitude Change During}} the {{Persian Gulf War}}} {A {{Coherence Model}} of {{Cognitive Consistency}}: {{Dynamics}} of {{Attitude Change During}} the {{Persian Gulf War}}}.{\BBCQ}
\newblock
\APACjournalVolNumPages{Journal of Social Issues}{49}{4}{147--165}.
\newblock
\begin{APACrefDOI} \doi{10.1111/j.1540-4560.1993.tb01185.x} \end{APACrefDOI}
\PrintBackRefs{\CurrentBib}

\bibitem [\protect \citeauthoryear {%
Sun%
\ \protect \BOthers {.}}{%
Sun%
\ \protect \BOthers {.}}{%
{\protect \APACyear {2024}}%
}]{%
SunEtAl2024Random}
\APACinsertmetastar {%
SunEtAl2024Random}%
\begin{APACrefauthors}%
Sun, S.%
, Lee, E.%
, Nan, D.%
, Zhao, X.%
, Lee, W.%
, Jansen, B\BPBI J.%
\BCBL {}\ \BBA {} Kim, J\BPBI H.%
\end{APACrefauthors}%
\unskip\
\newblock
\APACrefYearMonthDay{2024}{{\APACmonth{02}}}{}.
\newblock
\APACrefbtitle {Random {{Silicon Sampling}}: {{Simulating Human Sub-Population Opinion Using}} a {{Large Language Model Based}} on {{Group-Level Demographic Information}}} {Random {{Silicon Sampling}}: {{Simulating Human Sub-Population Opinion Using}} a {{Large Language Model Based}} on {{Group-Level Demographic Information}}}\ (\BNUM\ arXiv:2402.18144).
\newblock
\APACaddressPublisher{}{arXiv}.
\newblock
\begin{APACrefDOI} \doi{10.48550/arXiv.2402.18144} \end{APACrefDOI}
\PrintBackRefs{\CurrentBib}

\bibitem [\protect \citeauthoryear {%
Tang%
, Sun%
, Curran%
, Schaub%
\BCBL {}\ \BBA {} Shin%
}{%
Tang%
\ \protect \BOthers {.}}{%
{\protect \APACyear {2024}}%
{\protect \APACexlab {{\protect \BCnt {1}}}}}]{%
tang2024genai}
\APACinsertmetastar {%
tang2024genai}%
\begin{APACrefauthors}%
Tang, B\BPBI J.%
, Sun, K.%
, Curran, N\BPBI T.%
, Schaub, F.%
\BCBL {}\ \BBA {} Shin, K\BPBI G.%
\end{APACrefauthors}%
\unskip\
\newblock
\APACrefYearMonthDay{2024{\protect \BCnt {1}}}{}{}.
\newblock
{\BBOQ}\APACrefatitle {GenAI Advertising: Risks of Personalizing Ads with LLMs} {Genai advertising: Risks of personalizing ads with llms}.{\BBCQ}
\newblock
\APACjournalVolNumPages{arXiv}{2409.15436}{}{}.
\newblock
\begin{APACrefDOI} \doi{10.48550/arXiv.2409.15436} \end{APACrefDOI}
\PrintBackRefs{\CurrentBib}

\bibitem [\protect \citeauthoryear {%
Tang%
, Sun%
, Curran%
, Schaub%
\BCBL {}\ \BBA {} Shin%
}{%
Tang%
\ \protect \BOthers {.}}{%
{\protect \APACyear {2024}}%
{\protect \APACexlab {{\protect \BCnt {2}}}}}]{%
TangEtAl2024GenAI}
\APACinsertmetastar {%
TangEtAl2024GenAI}%
\begin{APACrefauthors}%
Tang, B\BPBI J.%
, Sun, K.%
, Curran, N\BPBI T.%
, Schaub, F.%
\BCBL {}\ \BBA {} Shin, K\BPBI G.%
\end{APACrefauthors}%
\unskip\
\newblock
\APACrefYearMonthDay{2024{\protect \BCnt {2}}}{{\APACmonth{09}}}{}.
\newblock
\APACrefbtitle {{{GenAI Advertising}}: {{Risks}} of {{Personalizing Ads}} with {{LLMs}}} {{{GenAI Advertising}}: {{Risks}} of {{Personalizing Ads}} with {{LLMs}}}\ (\BNUM\ arXiv:2409.15436).
\newblock
\APACaddressPublisher{}{arXiv}.
\newblock
\begin{APACrefDOI} \doi{10.48550/arXiv.2409.15436} \end{APACrefDOI}
\PrintBackRefs{\CurrentBib}

\bibitem [\protect \citeauthoryear {%
Tessler%
\ \protect \BOthers {.}}{%
Tessler%
\ \protect \BOthers {.}}{%
{\protect \APACyear {2024}}%
}]{%
TesslerEtAl2024AI}
\APACinsertmetastar {%
TesslerEtAl2024AI}%
\begin{APACrefauthors}%
Tessler, M\BPBI H.%
, Bakker, M\BPBI A.%
, Jarrett, D.%
, Sheahan, H.%
, Chadwick, M\BPBI J.%
, Koster, R.%
\BDBL {}Summerfield, C.%
\end{APACrefauthors}%
\unskip\
\newblock
\APACrefYearMonthDay{2024}{}{}.
\newblock
{\BBOQ}\APACrefatitle {AI can help humans find common ground in democratic deliberation} {Ai can help humans find common ground in democratic deliberation}.{\BBCQ}
\newblock
\APACjournalVolNumPages{Science}{386}{6719}{eadq2852}.
\newblock
\begin{APACrefDOI} \doi{10.1126/science.adq2852} \end{APACrefDOI}
\PrintBackRefs{\CurrentBib}

\bibitem [\protect \citeauthoryear {%
Thagard%
}{%
Thagard%
}{%
{\protect \APACyear {1989}}%
}]{%
thagardExplanatoryCoherence1989}
\APACinsertmetastar {%
thagardExplanatoryCoherence1989}%
\begin{APACrefauthors}%
Thagard, P.%
\end{APACrefauthors}%
\unskip\
\newblock
\APACrefYearMonthDay{1989}{{\APACmonth{09}}}{}.
\newblock
{\BBOQ}\APACrefatitle {Explanatory Coherence} {Explanatory coherence}.{\BBCQ}
\newblock
\APACjournalVolNumPages{Behavioral and Brain Sciences}{12}{3}{435--467}.
\newblock
\begin{APACrefDOI} \doi{10.1017/S0140525X00057046} \end{APACrefDOI}
\PrintBackRefs{\CurrentBib}

\bibitem [\protect \citeauthoryear {%
Wei%
\ \protect \BOthers {.}}{%
Wei%
\ \protect \BOthers {.}}{%
{\protect \APACyear {2023}}%
}]{%
wei2023chain}
\APACinsertmetastar {%
wei2023chain}%
\begin{APACrefauthors}%
Wei, J.%
, Wang, X.%
, Schuurmans, D.%
, Bosma, M.%
, Ichter, B.%
, Xia, F.%
\BDBL {}Zhou, D.%
\end{APACrefauthors}%
\unskip\
\newblock
\APACrefYearMonthDay{2023}{}{}.
\newblock
{\BBOQ}\APACrefatitle {Chain-of-Thought Prompting Elicits Reasoning in Large Language Models} {Chain-of-thought prompting elicits reasoning in large language models}.{\BBCQ}
\newblock
\APACjournalVolNumPages{arXiv preprint arXiv:2201.11903}{}{}{}.
\newblock
\begin{APACrefURL} \url{https://doi.org/10.48550/arXiv.2201.11903} \end{APACrefURL}
\newblock
\begin{APACrefDOI} \doi{10.48550/arXiv.2201.11903} \end{APACrefDOI}
\PrintBackRefs{\CurrentBib}

\bibitem [\protect \citeauthoryear {%
Weisman%
\ \BBA {} Markman%
}{%
Weisman%
\ \BBA {} Markman%
}{%
{\protect \APACyear {2017}}%
}]{%
weismanTheorybasedExplanationIntervention2017}
\APACinsertmetastar {%
weismanTheorybasedExplanationIntervention2017}%
\begin{APACrefauthors}%
Weisman, K.%
\BCBT {}\ \BBA {} Markman, E\BPBI M.%
\end{APACrefauthors}%
\unskip\
\newblock
\APACrefYearMonthDay{2017}{{\APACmonth{10}}}{}.
\newblock
{\BBOQ}\APACrefatitle {Theory-Based Explanation as Intervention} {Theory-based explanation as intervention}.{\BBCQ}
\newblock
\APACjournalVolNumPages{Psychonomic Bulletin \& Review}{24}{5}{1555--1562}.
\newblock
\begin{APACrefDOI} \doi{10.3758/s13423-016-1207-2} \end{APACrefDOI}
\PrintBackRefs{\CurrentBib}

\bibitem [\protect \citeauthoryear {%
Wilson%
\ \BBA {} Patterson%
}{%
Wilson%
\ \BBA {} Patterson%
}{%
{\protect \APACyear {1968}}%
}]{%
Wilson1968}
\APACinsertmetastar {%
Wilson1968}%
\begin{APACrefauthors}%
Wilson, G\BPBI D.%
\BCBT {}\ \BBA {} Patterson, J\BPBI R.%
\end{APACrefauthors}%
\unskip\
\newblock
\APACrefYearMonthDay{1968}{}{}.
\newblock
{\BBOQ}\APACrefatitle {A New Measure of Conservatism} {A new measure of conservatism}.{\BBCQ}
\newblock
\APACjournalVolNumPages{British Journal of Social and Clinical Psychology}{8}{}{264--269}.
\PrintBackRefs{\CurrentBib}

\end{thebibliography}

\end{document}